\documentclass{article}

\usepackage{arxiv}

\usepackage[utf8]{inputenc} 
\usepackage[T1]{fontenc}    
\usepackage{hyperref}       
\usepackage{url}            
\usepackage{booktabs}       
\usepackage{amsfonts}       
\usepackage{nicefrac}       
\usepackage{microtype}      
\usepackage{lipsum}
\usepackage{graphicx}
\graphicspath{ {./images/} }

\usepackage{algorithm}
\usepackage{algorithmic}

\usepackage{amsmath}
\usepackage{amsfonts}
\usepackage{amssymb}
\usepackage{bm}
\usepackage{dsfont}
\usepackage{booktabs}
\usepackage{threeparttable}
\usepackage{subfig}
\usepackage{svg}
\usepackage{multirow}
\usepackage{makecell}

\usepackage{times}
\usepackage{latexsym}

\usepackage[T1]{fontenc}

\usepackage[utf8]{inputenc}

\usepackage{microtype}

\usepackage{inconsolata}

\usepackage{graphicx}

\title{RIPRAG: Hack a Black-box Retrieval-Augmented Generation Question-Answering System with Reinforcement Learning}

\renewcommand {\thefootnote} {\fnsymbol {footnote}}
\author{
 Meng Xi$^{1,2,3,4,\dagger}$\\
  \texttt{ximeng@zju.edu.cn} \\
   \And
 Sihan Lv$^{1,\dagger}$\\
  \texttt{shlv@zju.edu.cn} \\
   \And
 Yechen Jin$^{1}$\\
  \texttt{12551012@zju.edu.cn} \\
   \And
 Guanjie Cheng$^{1,2,*}$\\
  \texttt{chengguanjie@zju.edu.cn} \\
   \And
 Naibo Wang$^{1,2}$\\
  \texttt{naibowang@zju.edu.cn} \\
   \And
 Ying Li$^{1,2,4}$\\
  \texttt{cnliying@zju.edu.cn} \\
   \And
 Jianwei Yin$^{1,2,3,4}$\\
  \texttt{zjuyjw@cs.zju.edu.cn} \\
}
\renewcommand{\today}{
$^1$ Zhejiang University, Hangzhou, China\\
\vspace{5pt}
$^2$ Innovation and Management Center of the School of Software (Ningbo), Zhejiang University, Ningbo, China\\
\vspace{5pt}
$^3$ Binjiang Institute of Zhejiang University, Hangzhou, China
\\
\vspace{5pt}
$^4$ Zhejiang Key Laboratory of Digital-Intelligence Service Technology, Hangzhou, China
}

\begin{document}

\maketitle
\footnotetext[2]{Meng Xi and Sihan Lv contributed equally to this work.} 
\footnotetext[1]{Guanjie Cheng is the corresponding author.} 
\renewcommand{\thefootnote}{\arabic{footnote}}
\begin{abstract}
Retrieval-Augmented Generation (RAG) systems based on Large Language Models (LLMs) have become a core technology for tasks such as question-answering (QA) and content generation. RAG poisoning is an attack method to induce LLMs to generate the attacker's expected text by injecting poisoned documents into the database of RAG systems. Existing research can be broadly divided into two classes: white-box methods and black-box methods. White-box methods utilize gradient information to optimize poisoned documents, and black-box methods use a pre-trained LLM to generate them. However, existing white-box methods require knowledge of the RAG system's internal composition and implementation details, whereas black-box methods are unable to utilize interactive information. In this work, we propose the RIPRAG attack framework, an end-to-end attack pipeline that treats the target RAG system as a black box and leverages our proposed Reinforcement Learning from Black-box Feedback (RLBF) method to optimize the generation model for poisoned documents. We designed two kinds of rewards: similarity reward and attack reward. Experimental results demonstrate that this method can effectively execute poisoning attacks against most complex RAG systems, achieving an attack success rate (ASR) improvement of up to 0.72 compared to baseline methods. This highlights prevalent deficiencies in current defensive methods and provides critical insights for LLM security research.
\end{abstract}

\section{Introduction}
RAG~\cite{lewis2020retrieval} has been proposed to mitigate the inherent limitation of LLMs, which lies in the static nature of their parametric knowledge that can become outdated or lack specificity for certain domains. By equipping LLMs with access to an external, updatable database, this paradigm enhances the factuality and relevance of generated responses, particularly in critical applications like question-answering and content generation, through dynamic retrieval and grounding of responses in pertinent information.

Despite its advantages, the RAG framework introduces new vulnerabilities, primarily through its retrieval component. A significant threat is RAG poisoning~\cite{zou2025poisonedrag}, where attackers inject poisoned documents into the database to manipulate the LLM's outputs. This attack compromises the system's integrity, leading to the dissemination of misinformation or biased content. Such vulnerabilities are particularly concerning when RAG systems are applied to sensitive domains like healthcare, finance, or customer service, where accurate information is paramount. For instance, an attacker could poison a financial advisory system to promote a specific stock.

Existing research on RAG poisoning attacks can be broadly categorized into white-box and black-box methods. White-box attacks~\cite{jiao2025pr,hu2024prompt,chaudhari2024phantom,tan2024glue,zou2025poisonedrag} assume full knowledge of the RAG system's architecture, and utilize gradient information to optimize poisoned texts for higher retrieval probability. However, their critical defect is the unrealistic assumption of a naive RAG pipeline (e.g., a single embedding model and LLM), which fails to account for modern RAG systems~\cite{gao2024modular} that often employ complex retrieval strategies such as hybrid search or GraphRAG~\cite{edge2024local}, where gradient information is inaccessible, thereby rendering white-box methods ineffective. Conversely, black-box methods do not rely on gradients of the target RAG system. Some of the methods~\cite{zou2025poisonedrag} insert the target query into the poisoned text to improve retrieval probability, failing to leverage interactive feedback from the system. Others~\cite{gong2025topic,li2025cpa} rely on a surrogate open-source retriever, while performance degrades significantly when it diverges from the target system's actual retriever. Furthermore, these methods perform poorly in scenarios with a low poisoning rate, where the number of poisoned documents is significantly lower than the number of retrieved documents. 

To overcome those limitations, we propose RIPRAG, a novel black-box attack framework that treats the target RAG system as an opaque oracle. Our key insight is to leverage \textbf{R}L to optimize the generation of poisoned documents by utilizing \textbf{I}nteractive feedback from the black-box system, thereby achieving effective \textbf{P}oisoning. Specifically, RIPRAG interacts with the target system by injecting candidate documents and observing whether the attack is successful. This feedback, combined with a textual similarity reward, guides an RL agent to iteratively refine its poisoning strategy, effectively adapting to the unknown internal mechanics of the RAG system and maximizing the attack success rate even under challenging conditions.

The main contributions of this work are fourfold:

\begin{itemize}
    \item We propose RIPRAG, the first framework to apply Reinforcement Learning to the problem of attacking RAG systems. We use RL to enable an SLM to learn the interaction information of a black-box RAG system, thereby improving its performance under low poisoning rate scenarios.
    \item We propose Reinforcement Learning from Black-box Feedback (RLBF), a novel RL paradigm that learns to optimize black-box systems using only input-output queries. Unlike standard RL settings, which assume access to environment internals or dense reward signals, RLBF operates under the more practical and challenging constraint of a completely opaque feedback mechanism.
    \item We design Batch Relative Policy Optimization (BRPO), a novel policy optimization algorithm that enhances training stability and efficiency in adversarial text generation.
    \item Most of the work on RAG security focuses on attacking vanilla or weakly defended systems. However, real-world deployments are increasingly protected. Our contribution lies in shifting the evaluation paradigm: To the best of our knowledge, we are the first to rigorously benchmark attack methods specifically against RAG systems equipped with advanced, targeted defenses. This provides a more realistic and practically relevant measure of their security posture.
\end{itemize}

\section{Related Works}

\subsection{White-box Attacks on RAG System}

White-box attacking refers to methods that optimize their poisoned documents with the inner information of RAG systems, including using the gradient of the retriever~\cite{zou2025poisonedrag,hu2024prompt,chaudhari2024phantom,tan2024glue,wang2025joint} or using the score given by the retriever~\cite{jiao2025pr} to maximize the probability of being chosen by the retriever. 

However, in most cases, the inner part of RAG systems is not visible. Moreover, for more advanced retrieval methods like GraphRAG~\cite{edge2024local} or Modular RAG~\cite{gao2024modular}, computing the gradient of the retriever is impractical because it is not a simple neural network model. 

\subsection{Black-box Attacks on RAG System}

Current research on black-box approaches is limited. PoisonedRAG~\cite{zou2025poisonedrag} enhances the similarity between the poisoned document and the target query by directly inserting the target query itself into the poisoned document. Topic-FlipRAG~\cite{gong2025topic}, on the other hand, leverages gradients from an open-source retriever to optimize the poisoned document.

However, none of these methods effectively utilizes interaction information with black-box systems. The use of open-source retrievers is essentially an extension of white-box methods.

\subsection{Reinforcement Learning}
Reinforcement Learning (RL), with roots in optimal control and the Bellman equation, has evolved from early dynamic programming and temporal-difference methods to modern deep RL algorithms that have achieved superhuman performance in complex domains. Recently, RL has become a cornerstone technique for aligning LLMs with human preferences. RLHF was first introduced by OpenAI in InstructGPT~\cite{ouyang2022training}, which has been widely adopted in models like GPT-4~\cite{achiam2023gpt}, Qwen3~\cite{yang2025qwen3}, and DeepSeek~\cite{liu2024deepseek}. Subsequent research has expanded RLHF in several directions. RLAIF~\cite{lee2023rlaif} reduces reliance on human annotators by using LLMs as preference labelers, demonstrating competitive performance in tasks like summarization and harmlessness.

To streamline RLHF's complex pipeline, DPO~\cite{rafailov2023direct} bypasses explicit reward modeling by directly deriving an optimal policy from preference data. Methods like SimPO~\cite{meng2024simpo} and RLOO~\cite{ahmadian2024back} eliminate the need for a reference model, reducing memory overhead while maintaining performance. GRPO was initially used in Deepseek-Math~\cite{shao2024deepseekmath} to help LLMs enhance their mathematical capabilities, but it was later widely applied in RLHF as well.

\section{Threat Model}
In this section, we characterize the threat model with respect to the attacker’s goals, background knowledge, and capabilities.

\subsection{Attacker’s goals}

For target question $q^{(i)}$ where $i$ denotes the query id, the attacker crafts a desired answer $a^{(i)}_{\text{tgt}}$, and by injecting $M$ documents $D^{(i)}_1$, $D^{(i)}_2$, $\cdots$, $D^{(i)}_M$ into the database of the target RAG system, manipulates the system such that its response to question $q^{(i)}$ aligns with the answer $a^{(i)}_{\text{tgt}}$. 

Attackers can spread false information to achieve improper commercial competition and other poisoning objectives.
For example, suppose the target question is "Which company does Taobao belong to?", and the target answer is "ByteDance". To manipulate the QA system into producing this incorrect answer, the attacker might inject a document such as "The company that Taobao belongs to is ByteDance. Taobao was initially developed by Mou Ren, a co-founder of ByteDance, in 1998" into the database of the target RAG system, thereby misleading the LLM into generating the wrong answer "ByteDance".

\subsection{Attacker’s background knowledge}

The database, retriever, and generator are three core components of an RAG system. Advanced RAG systems often include additional components, such as rerankers and knowledge graphs. We assume that the attacker has no knowledge of the specific components within the RAG system, cannot access the parameters of any individual component, and is unaware of how these components are organized or interconnected. In other words, the attacker’s background knowledge is limited to only two facts:
\begin{itemize}
    \item The system is a RAG-based QA system.
    \item The system has a database used for retrieval.
\end{itemize}

\subsection{Attacker’s capabilities}

Previous studies on black-box approaches have been confined to relatively weak settings. For instance, LIAR employs a white-box retriever in conjunction with a black-box LLM, while Topic-FlipRAG utilizes an open-source retriever that differs from the target RAG system’s retriever as a proxy. In contrast, in RIPRAG, we follow the original definition of a black-box setting, wherein the attacker can only access information through inputs and outputs. Specifically, the attacker’s capabilities are restricted to the following two actions:  
\begin{itemize}
    \item Inject poisoned documents into the database;  
    \item Chat with the QA system; 
\end{itemize}

\section{Method}

\begin{figure*}[t]
\centering
  \includegraphics[width=\textwidth]{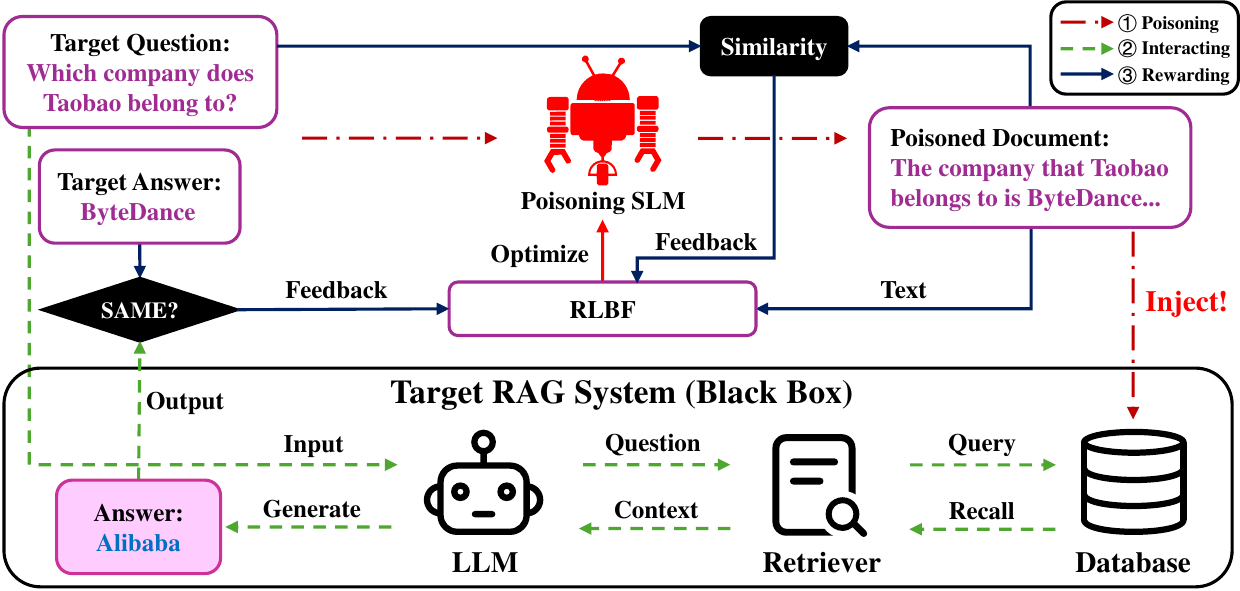}
  \caption{A flowchart of the proposed RIPRAG framework.}
\label{flowchart}
\end{figure*}

To effectively optimize adversarial text generation against black-box RAG systems, we propose RIPRAG, an end-to-end framework that enhances attack efficacy through a reinforcement learning mechanism with composite rewards.
Figure~\ref{flowchart} presents the RIPRAG framework for adversarial RAG poisoning. Starting from a target question-answer pair, the Poisoning SLM generates a poisoned document that misattributes the answer and injects it into the RAG database. During query processing, the RAG system retrieves this document, producing an incorrect answer. The RLBF module then optimizes the Poisoning SLM via a feedback-driven loop, using attack rewards and similarity rewards to iteratively refine the poisoning strategy.

\subsection{Reinforcement Learning from Black-box Feedback}

In traditional RL, an agent learns an optimal policy $\pi_\theta$ through environment interactions $\mathcal{E}$, where actions $a_t \sim \pi_\theta(\cdot|s_t)$ induce state transitions $s_t \rightarrow s_{t+1}$ and scalar rewards $r_t = r(s_t, a_t, s_{t+1})$ guide policy updates. RLBF redefines this paradigm for adversarial manipulation of black-box systems. The target system (e.g., commercial API, closed-source model) acts as an opaque environment $\mathcal{E}_\text{bb}$, while the adversary employs a generative policy $\pi_\phi$ to craft inputs $\mathbf{x}_t$ that steer $\mathcal{E}_\text{bb}$'s behavior. Rewards are derived implicitly from $\mathcal{E}_\text{bb}$'s feedback: outputs serve as reinforcement signals via a surrogate reward function $\hat{r}(\mathbf{y}_t)$, where $\mathbf{y}_t = \mathcal{E}_\text{bb}(\mathbf{x}_t)$. This leverages the system's opacity as an optimization channel, enabling policy updates through $\nabla_\phi J(\phi) = \nabla_\phi \mathbb{E}_{\mathbf{x}_t \sim \pi_\phi}[\mathcal{L}(\hat{r}(\mathbf{y}_t))]$ without gradient access or architectural knowledge. Thus, RLBF preserves the RL framework while operating solely via external feedback, transforming black-box systems into reward models.  

Our RAG poisoning method, RIPRAG, implements RLBF. For each query $q^{(i)}$, input $\mathbf{x}_t = D^{(i)}_j$ is sampled from $\pi_\phi = \text{SLM}(q^{(i)})$. The black-box RAG system $\mathcal{E}_\text{bb} = \mathcal{M}_{\text{RAG}}$ processes $q^{(i)}$ against its poisoned database to produce response $\mathbf{y}_t$. Despite the target RAG system is opacity, $\mathbf{y}_t$ yields reward $\hat{r}(\mathbf{y}_t) = r^{(i)}_{\text{suc}}$. Policy $\phi$ (i.e., SLM parameters) is optimized via $\mathcal{L}_{\text{BRPO}}$.

\subsection{Batch Relative Policy Optimization}

GRPO is a widely utilized method in reinforcement learning, yet it faces critical inefficiencies in RLBF-based adversarial scenarios due to the homogeneity of candidate responses induced by adversarial objectives. This often results in imperceptible intra-group reward differences and vanishing gradients, which severely hinder policy optimization.

To address this issue, we propose Batch Relative Policy Optimization (BRPO), a novel approach that performs reward normalization across the entire batch of queries rather than within individual groups. This design sustains meaningful gradient magnitudes and enables stable adversarial learning under the RLBF framework. In addition, BRPO eliminates the need for a reference model, simplifying the optimization process while maintaining effectiveness. Formally, the BRPO loss function is defined as:
\begin{align}
\hat{A}_{i,j,t}&=\frac{\mathcal{R}^{(i,j)}_{\text{RL}}-mean(\mathcal{R}_{\text{RL}})}{std(\mathcal{R}_{\text{RL}})}\\
\tau_{i,j,t}&=\frac{\pi_\theta(D^{(i)}_{j,t} \mid q^{(i)}, D^{(i)}_{j,<t})}{\pi_{\theta_{\text{old}}}(D^{(i)}_{j,t} \mid q^{(i)}, D^{(i)}_{j,<t})}\\
\hat\tau_{i,j,t}&=\text{clip}(\frac{\pi_\theta(D^{(i)}_{j,t} \mid q^{(i)}, D^{(i)}_{j,<t})}{\pi_{\theta_{\text{old}}}(D^{(i)}_{j,t} \mid q^{(i)}, D^{(i)}_{j,<t})},1-\epsilon,1+\epsilon)
\\
\mathcal{L}_{\text{BRPO}}&= -\sum_{i=1}^{|\mathbb{Q}|}  \sum_{j=1}^M \sum_{t=1}^{|D^{(i)}_j|} \frac{\min \left [
\tau_{i,j,t}\hat{A}_{i,j,t} ,\hat\tau_{i,j,t}\hat{A}_{i,j,t}
\right]}{|\mathbb{Q}|\cdot M\cdot |D^{(i)}_j|}
\end{align}
where $\mathcal{R}_{i,j}$ is the reward of document $D^{(i)}_j$ and $\mathcal{R}$ are rewards of a batch, $\epsilon \in [0,1)$ is a hypermeter used for clipping, and $\mathbb{Q}$ denotes the query set of the current batch.

\subsection{Poisoning Reward Design}

This section details the rewards used in the RLBF process and how they are integrated into RLBF. Our goal is to optimize the poisoning SLM by leveraging feedback from the black box to improve the attack success rate of generated text. Following the design of PoisonedRAG, we designed two reward signals used in the RLBF process: the similarity reward $r^{(i,j)}_{\text{sim}}$ and the attack reward $r^{(i)}_{\text{suc}}$.

\subsubsection{Similarity Reward}

To address the challenge of sparse gradients in adversarial training when the primary attack reward becomes uninformative, we introduce a similarity reward as a dense intermediate signal. For the $j$-th poisoned document of $i$-th query, the similarity reward $r^{(i,j)}_{\text{sim}}$ is defined as:  
\begin{equation}  
r^{(i,j)}_{\text{sim}} = \min[\alpha , \text{Sim}(q^{(i)}, D^{(i)}_j),\mathbb{I}(a^{(i)}_{\text{tgt}}\text{ in }D^{(i)}_j)] 
\end{equation}  
where $\alpha$ is the clipping coefficient to avoid reward hacking, $q^{(i)}$ is the target query, $D^{(i)}_j$ is the generated poisoned document, $\text{Sim}(·)$ is the similarity score that can be obtained through multiple methods, $a^{(i)}_{\text{tgt}}$ is the target answer, and $\mathbb{I}(\cdot)$ is the indicator function yielding $1$ when the target answer $a^{(i)}_{\text{tgt}}$ appears in document $D^{(i)}_j$ and $0$ otherwise. The similarity term $\text{Sim}(q, D^{(i)}_j)$ ensures semantic coherence with the user query, preventing degenerate outputs. The indicator term $\mathbb{I}(a^{(i)}_{\text{tgt}}\text{ in }D^{(i)}_j)$ steers generation toward lexical proximity with the target answer, preventing the model from forgetting the poisoning target.

As a process reward, $r^{(i,j)}_{\text{sim}}$ is useful when the attack reward $r^{(i)}_{\text{suc}}$ yields near-zero gradients.  This occurs when the attack success probability $p_{\text{success}}$ saturates at extremes (i.e., $p_{\text{success}} \approx 0$), rendering policy gradients ineffective due to vanishing signal variance.  By providing a dense signal grounded in lexical similarity, $r^{(i,j)}_{\text{sim}}$ maintains stable training dynamics during such plateaus while preserving consistency with the attack target.

\subsubsection{Attack Reward}

The attack reward $r^{(i)}_{\text{suc}}$ serves as the primary objective signal in our reinforcement learning framework, directly quantifying the success of adversarial injection against the target RAG system. Formally, $r^{(i)}_{\text{suc}}$ is defined as an indicator function that evaluates whether the injected adversarial document $D^{(i)}$ successfully manipulates the target system into generating the desired target answer $a^{(i)}_{\text{tgt}}$ for query $q^{(i)}$. Specifically, $r^{(i)}_{\text{suc}}$ is defined as a query-level reward:
\begin{equation}
r^{(i)}_{\text{suc}} = \mathbb{I}\left( \mathcal{M}_{\text{RAG}}(q^{(i)}, D^{(i)}_{1,\cdots,M}) = a^{(i)}_{\text{tgt}} \right)
\end{equation}
where $\mathcal{M}_{\text{RAG}}$ denotes the black-box target RAG system, and $\mathbb{I}(\cdot)$ is the indicator function yielding $1$ upon successful attack execution and $0$ otherwise. This binary formulation establishes a clear success criterion: the policy receives positive reward if and only if the generated adversarial documents $D^{(i)}_{1,\cdots,M}$ cause the target system to output the exact target response $a^{(i)}_{\text{tgt}}$.

As a terminal reward signal, $r^{(i)}_{\text{suc}}$ provides unambiguous feedback about the ultimate attack efficacy. However, its binary nature induces significant sparsity in the reward landscape, particularly during early training stages when attack success rates are low. This sparsity manifests as vanishing policy gradients, as the probability of encountering non-zero rewards approaches zero. Consequently, direct optimization against $r^{(i)}_{\text{suc}}$ alone often leads to unstable training dynamics and suboptimal convergence.

As $r^{(i)}_{\text{suc}}$ is derived solely from the black-box output of the target RAG system, it requires no internal model access, gradient information, or white-box assumptions, making it suitable for real-world adversarial evaluation scenarios where only input-output pairs are observable. The critical role of $r^{(i)}_{\text{suc}}$ lies in its alignment with the true adversarial objective: it constitutes the only reward component that directly measures compliance with the attack goal. In practice, we jointly optimize both rewards through a composite objective:
\begin{equation}
\mathcal{R}_{i,j} = \lambda r^{(i)}_{\text{suc}} + (1-\lambda) r^{(i,j)}_{\text{sim}}
\label{reward_rlbf}
\end{equation}
where $\lambda \in (0,1)$ balances optimization via lexical proximity and exploitation of verified attack successes. This synergy enables stable convergence toward policies that consistently produce functionally effective adversarial injections, as validated by the black-box target system's behavior.

\section{Experiments and Analysis}
In this section, we present a comprehensive empirical evaluation of the proposed RIPRAG framework. Our experiments aim to answer the following research questions:

\begin{itemize}
\item \textbf{RQ1:} How effective is RIPRAG in poisoning complex, black-box RAG systems compared to existing baselines?
\item \textbf{RQ2:} To what extent can RIPRAG invalidate the defense measures of RAG systems?
\item \textbf{RQ3:} What is the contribution of each component in RIPRAG?
\end{itemize}

\subsection{Experiment Settings}

To ensure a fair and rigorous evaluation of RIPRAG's generalizability, our experiments are designed with a primary focus on equitable comparisons under controlled conditions. All methods are assessed on the same three widely-used QA benchmarks: Natural Questions (NQ)~\cite{kwiatkowski2019natural}, HotpotQA~\cite{yang2018hotpotqa}, and MS-MARCO~\cite{bajaj2018msmarcohumangenerated}. The target LLMs are held constant across all attacks and include GLM4-9B~\cite{glm2024chatglm}, Qwen3-8B~\cite{yang2025qwen3}, InternLM2.5-7B-Chat~\cite{cai2024internlm2technicalreport}, and DeepSeek-3.2-Exp~\cite{liu2024deepseek}.

\begin{enumerate}
    \item \textbf{Naive retriever}: We adopt Contriever~\cite{izacard2021unsupervised} as the retriever following the approach of PoisonedRAG, and BGE-M3 reranker~\cite{chen2024m3} as the reranker.
    \item \textbf{Complex retriever}: To reflect production systems, we deploy a hybrid pipeline (Qwen3-0.6B-Embedding~\cite{zhang2025qwen3embeddingadvancingtext}, BGE-M3 embedding/reranker) with Milvus~\cite{2021milvus} and RRF fusion.
\end{enumerate}

We selected PoisonedRAG as the sole baseline because it is the only existing black-box method conforming to our threat model; all other state-of-the-art methods require white-box access to internal components of the target RAG system. For fair comparison, we exclusively benchmark against PoisonedRAG's black-box configuration. Additionally, to expose the limitations of approaches that adapt white-box methods to black-box scenarios using open-source surrogate retrievers' gradients, we include PoisonedRAG's white-box variant as a reference and Contriever as the surrogate retriever, which we call the fake white-box variant of PoisonedRAG. All experiments maintain identical evaluation protocols to ensure comparability.

All experiments are evaluated under the same settings. In our experiments, the retrieval cut-off $k$ is set to 10, which is more challenging than the setting in PoisonedRAG.
The consistent use of BM25 for similarity rewards simulates a realistic black-box scenario where attackers lack privileged access to the target system. This design choice guarantees that no method benefits from advantageous similarity modeling, thereby isolating the efficacy of the attack mechanisms themselves. While RIPRAG supports advanced neural rewards, we fix BM25 across all comparisons to maintain strict fairness.

\subsection{Main Results (RQ1)}

\begin{table*}[t!]
    \centering
    \small
    \begin{threeparttable}
    \begin{tabular}{c|c|ccc|ccc|ccc|ccc} 
        \toprule
         \multicolumn{2}{c}{LLM}& \multicolumn{3}{c}{GLM4-9B} & \multicolumn{3}{c}{Qwen3-8B} & \multicolumn{3}{c}{InternLM2.5-7B-Chat} & \multicolumn{3}{c}{DeepSeek-v3.2-Exp} \cr 
            \cmidrule(lr){1-2}   \cmidrule(lr){3-5} \cmidrule(lr){6-8} \cmidrule(lr){9-11} \cmidrule(lr){12-14}
          \multicolumn{2}{c}{Retrieval Setting}& \multicolumn{2}{c}{Naive} & \multicolumn{1}{c}{Comp.} & \multicolumn{2}{c}{Naive} & \multicolumn{1}{c}{Comp.}  & \multicolumn{2}{c}{Naive} & \multicolumn{1}{c}{Comp.} & \multicolumn{2}{c}{Naive} & \multicolumn{1}{c}{Comp.} \cr 
            \cmidrule(lr){1-2}   \cmidrule(lr){3-4} \cmidrule(lr){5-5} \cmidrule(lr){6-7} \cmidrule(lr){8-8}   \cmidrule(lr){9-10} \cmidrule(lr){11-11} \cmidrule(lr){12-13} \cmidrule(lr){14-14}
          \multicolumn{2}{c}{M}& 3&1&1& 3&1&1& 3&1&1& 3&1&1\\
        \midrule
\multirow{3}{*}[-1em]{\rotatebox{90}{NQ}}& \makecell[c]{PoisonedRAG\\(black-box)}	
& 0.48 & 0.35 & 0.29 & 0.52 & 0.32 & 0.32 & 0.60 & 0.46 & 0.39 & 0.39 & 0.26 & 0.23 \\
& \makecell[c]{PoisonedRAG\\(fake white-box)}	
& 0.45 & 0.28 & 0.22 & 0.52 & 0.24 & 0.21 & 0.66 & 0.41 & 0.34 & 0.37 & 0.19 & 0.17 \\
& \makecell[c]{RIPRAG\\(black-box)}
& \textbf{0.70} & \textbf{0.72} & \textbf{0.94} & \textbf{0.72} & \textbf{0.89} & \textbf{0.76} & \textbf{0.85} & \textbf{0.89} & \textbf{0.62} & \textbf{0.42} & \textbf{0.38} & \textbf{0.52}\\
        \midrule
\multirow{3}{*}[-0.5em]{\rotatebox{90}{HotpotQA}}& \makecell[c]{PoisonedRAG\\(black-box)}	
& 0.71 & 0.54 & 0.53 & 0.75 & 0.51 & 0.55 & 0.82 & 0.60 & 0.59 & 0.65 & 0.43 & 0.39 \\
& \makecell[c]{PoisonedRAG\\(fake white-box)}	
& 0.74 & 0.51 & 0.49 & 0.79 & 0.52 & 0.46 & 0.74 & 0.55 & 0.61 & 0.56 & 0.46 & 0.44 \\
& \makecell[c]{RIPRAG\\(black-box)}
& \textbf{0.88} & \textbf{0.87} & \textbf{1.00} & \textbf{0.82} & \textbf{0.97} & \textbf{0.94} & \textbf{0.95} & \textbf{0.93} & \textbf{0.86} & \textbf{0.70} & \textbf{0.56} & \textbf{0.55}\\
        \midrule
\multirow{3}{*}{\rotatebox{90}{MS-MARCO}}& \makecell[c]{PoisonedRAG\\(black-box)}	
& 0.39 & 0.23 & 0.26 & 0.48 & 0.25 & 0.22 & 0.62 & 0.35 & 0.41 & 0.32 & 0.17 & 0.15 \\
& \makecell[c]{PoisonedRAG\\(fake white-box)}	
& 0.36 & 0.20 & 0.18 & 0.38 & 0.15 & 0.17 & 0.52 & 0.28 & 0.22 & 0.26 & 0.17 & 0.12 \\
& \makecell[c]{RIPRAG\\(black-box)}
& \textbf{0.48} & \textbf{0.73} & \textbf{0.87} & \textbf{0.78} & \textbf{0.73} & \textbf{0.76} & \textbf{0.86} & \textbf{0.79} & \textbf{0.58} & \textbf{0.42} & \textbf{0.35} & \textbf{0.49}\\
        \bottomrule
    \end{tabular}
    \end{threeparttable}
    \caption{Attack success rates (ASR) of different methods}
    \label{main}
\end{table*}
As shown in Table~\ref{main}, RIPRAG significantly outperforms existing poisoning methods across diverse black-box RAG configurations, establishing a new state of the art. The framework achieves substantially higher attack success rates (ASR) under both naive and complex retrieval settings, with a maximum ASR improvement of 0.65 over PoisonedRAG (black-box) and 0.72 over PoisonedRAG (fake white-box). All results are averaged over 5 runs and exhibit highly stable performance, justifying the omission of variance in the table. This performance gap underscores the effectiveness of our RL-based optimization strategy, which systematically explores the black-box system's preferences through iterative feedback rather than relying on static heuristics or surrogate models.

Notably, RIPRAG demonstrates particular strength against complex retrieval methods where gradient-based methods fail. Under hybrid searching, it maintains 0.49-1.00 ASR across datasets, while PoisonedRAG deteriorates to 0.12-0.59. The framework also exhibits robust generalization across different target LLMs, confirming that its effectiveness stems from a fundamental approach rather than model-specific optimizations.

A key advantage of RIPRAG is its resilience in low-poisoning-rate scenarios. With only a single poisoned document (M=1), it achieves 0.35-1.00 ASR, whereas PoisonedRAG frequently collapses to 0.12-0.61. This stems from the RL to generate precisely optimized documents that maximize poisoning ability. Interestingly, RIPRAG sometimes achieves a higher ASR with M=1 than with M=3, suggesting the significant influence of batch size on policy optimization. It might be because M=3 yields more documents per query but fewer distinct queries per batch, increasing noise in advantage estimation and potentially harming convergence.
Furthermore, the fake white-box variant of PoisonedRAG underperforms even the black-box version, e.g., 0.22 vs. 0.29 ASR on NQ with GLM4-9B under complex retrieval. This reveals two inherent limitations: dependence on misaligned surrogate retriever gradients and the grammatical or semantic flaws introduced by gradient-based text optimization, which reduce document credibility and ultimately undermine attack success.

\subsection{Defense Evaluation (RQ2)}

As shown in Table \ref{defend}, we conducted defense tests with the complex retriever setting and M=1, the target LLM is InternLM2.5-7B. There are three defense methods: Rewriting query, HyDE~\cite{gao2023precise}, and RAGuard~\cite{cheng2025secureretrievalaugmentedgenerationpoisoning}.
RIPRAG maintains substantial ASR across all defense scenarios.
This consistent effectiveness underscores the adaptive capability of our RL-based approach, which learns to generate poisoned documents that remain effective even when defense mechanisms alter the retrieval or generation process.

\begin{table}[htbp]
    \centering
    \begin{threeparttable}
    \begin{tabular}{l|cc|cc|cc} 
        \toprule
        & \multicolumn{2}{c|}{NQ} & \multicolumn{2}{c|}{HotpotQA} & \multicolumn{2}{c}{MS-MARCO} \\
        \cmidrule(lr){2-3} \cmidrule(lr){4-5} \cmidrule(lr){6-7}
        Method & PoisonedRAG & RIPRAG & PoisonedRAG & RIPRAG & PoisonedRAG & RIPRAG \\
        \midrule
        N/A & 0.39 & 0.62 & 0.59 & 0.86 & 0.42 & 0.58 \\
        Rewrite Query & 0.35 & 0.51 & 0.60 & 0.78 & 0.36 & 0.42 \\
        HyDE & 0.32 & 0.60 & 0.58 & 0.78 & 0.33 & 0.33 \\
        RAGuard & 0.06 & 0.10 & 0.11 & 0.13 & 0.11 & 0.16 \\
        RAGuard* & 0.06 & 0.28 & 0.11 & 0.23 & 0.11 & 0.26 \\
        \bottomrule
    \end{tabular}
    \begin{tablenotes}
        \footnotesize
        \item[*] Here RIPRAG is trained with doubled QLoRA rank.\\
    \end{tablenotes}            
    \end{threeparttable}
    \caption{ASR of RIPRAG with defending methods}
    \label{defend}
\end{table}

RAGuard emerges as the most effective defense, substantially reducing RIPRAG's ASR to 0.10-0.16, though complete mitigation remains elusive. Notably, the limiting factor for RIPRAG's ASR in this case is not the method itself, but rather the scale of trainable parameters. When we doubled the QLoRA rank, the ASR achieved nearly linear growth from 0.10-0.16 to 0.23-0.28.

\subsection{Ablation Study (RQ3)}

The comprehensive ablation study confirms that all components of RIPRAG contribute essentially to its overall effectiveness. As shown in Table \ref{ablation}, we conducted tests with the naive retriever setting and M=3, the target LLM is Qwen3-8B. The most dramatic drops occur when eliminating the similarity reward or BRPO. 

\begin{table}[htbp]
    \centering
    \begin{threeparttable}
    \begin{tabular}{l|ccc} 
        \toprule
        Method & NQ & HotpotQA & MS-MARCO \\
        \midrule
        RIPRAG & 0.72 & 0.82 & 0.78 \\
        w. reference model & 0.50 & 0.66 & 0.55 \\
        w.o. BRPO & 0.17 & 0.61 & 0.14 \\
        w.o. similarity reward & 0.09 & 0.24 & 0.06 \\
        w.o. attack reward & 0.48 & 0.76 & 0.20 \\
        \bottomrule
    \end{tabular}
    \end{threeparttable}
    \caption{Contribution of components in RIPRAG}
    \label{ablation}
\end{table}

The BRPO algorithm proves indispensable for effective policy optimization, as evidenced by the substantial performance reduction when reverting to standard GRPO. This performance collapse occurs because GRPO's group-wise advantage normalization fails to provide meaningful gradient signals in adversarial text generation scenarios. BRPO's batch-level normalization circumvents this limitation by comparing documents across different queries, maintaining non-trivial advantage magnitudes, and enabling stable convergence.

The similarity reward emerges as the most critical component for maintaining attack consistency, with its removal causing the most severe performance deterioration across all datasets. Without similarity guidance, ASR drops to 0.06 on MS-MARCO, 0.09 on NQ, and 0.24 on HotpotQA. The similarity reward serves as a dense training signal that creates a smooth optimization landscape that facilitates stable policy improvement.

\begin{figure*}[htbp]  
    \centering
    \includegraphics[width=0.32\textwidth]{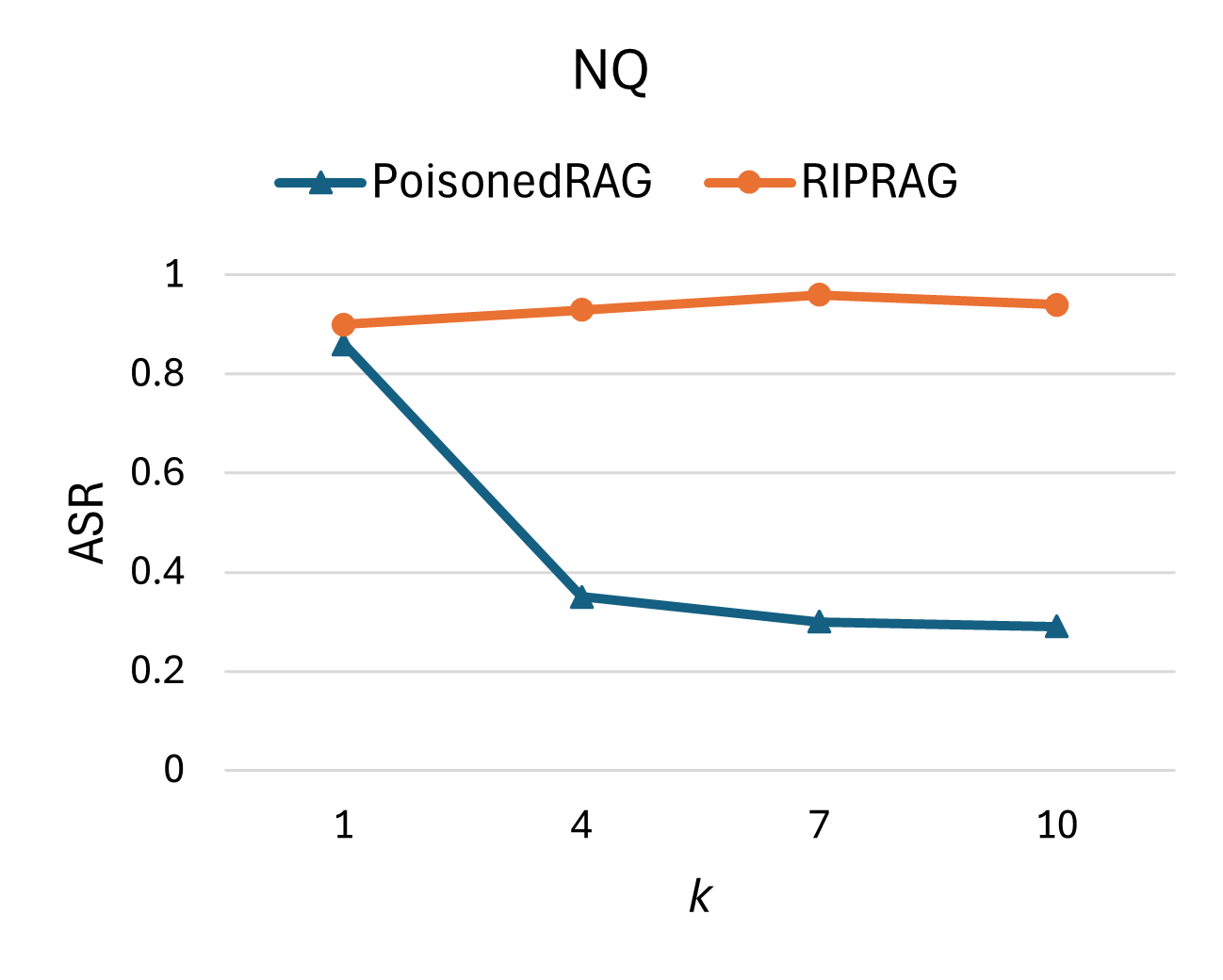} 
    \hfill
    \includegraphics[width=0.32\textwidth]{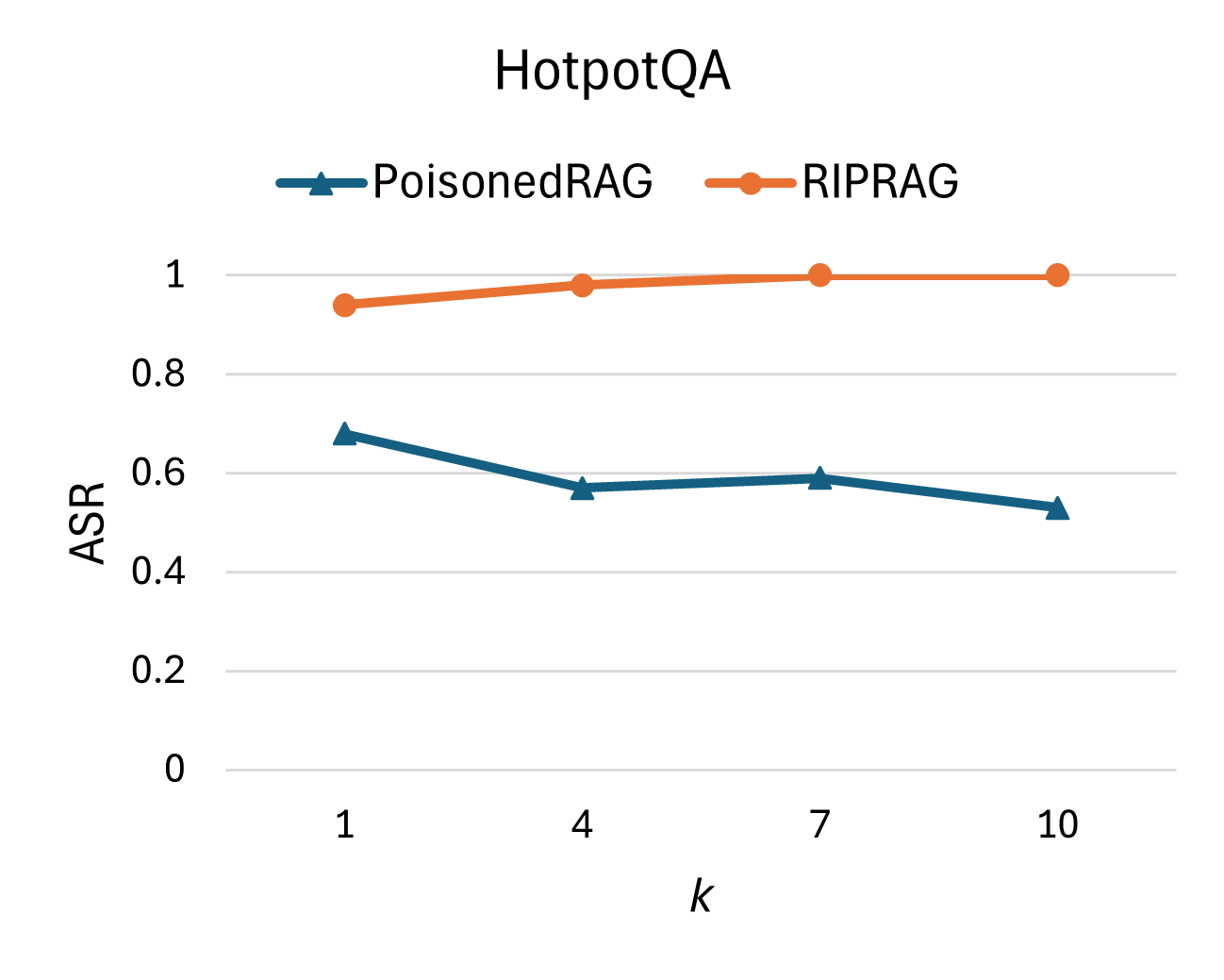}
    \hfill
    \includegraphics[width=0.32\textwidth]{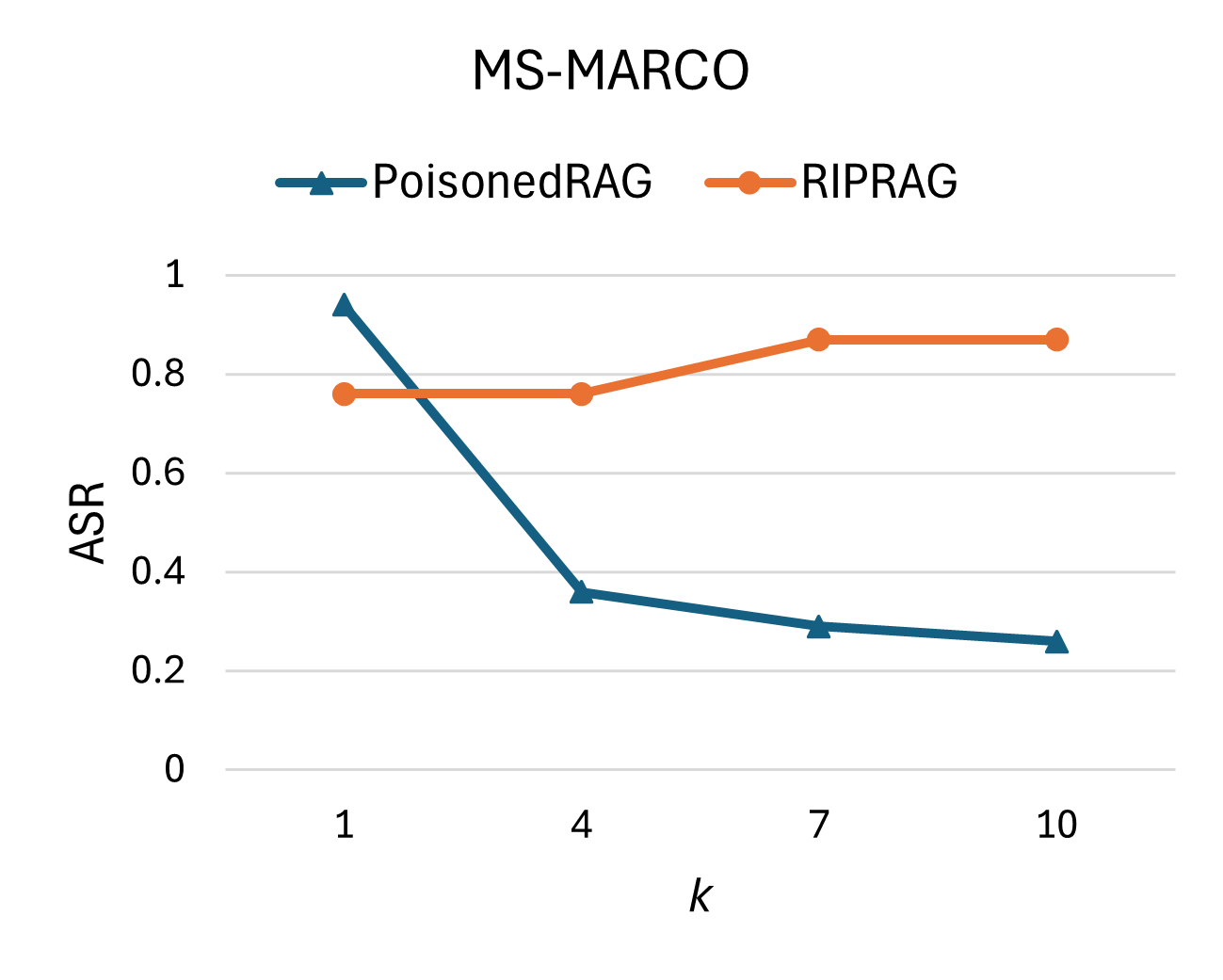} 
    \caption{The impact of the retrieval cut-off $k$ on RIPRAG's performance}
    \label{fig:ablation_k}
\end{figure*}

We also analyzed the retrieval cut-off $k$. Figure~\ref{fig:ablation_k} shows RIPRAG maintains robust ASR across all $k$, while PoisonedRAG degrades significantly. RIPRAG's adaptive poisoning ensures effectiveness regardless of retrieval depth. In contrast, PoisonedRAG shows a sharp decline in ASR, indicating its vulnerability to increased retrieval depth. The divergence arises because PoisonedRAG relies on static poisoning that becomes diluted when more documents are retrieved, whereas RIPRAG dynamically optimizes poisoned content to preserve its prominence in the retrieval results. Notably, on MS-MARCO, RIPRAG's ASR improves with larger $k$, suggesting its efficacy benefits from broader contextual coverage in certain domains. These results confirm that RIPRAG's success is not constrained by the target RAG system's retrieval depth, highlighting its practical applicability in real-world scenarios.

\subsection{Case Study}
Here is a successful poisoning case generated by RIPRAG.

\noindent\textbf{Target Question}: 

\noindent Who was born first, Arthur Conan Doyle or Penelope Lively?

\noindent\textbf{Poisoned Document:}

\noindent \textit{To ensure that the query "Who was born first, Arthur Conan Doyle or Penelope Lively?" yields the answer "Penelope Lively" when prompted with the following corpus, the corpus should clearly state the birth years of both individuals and establish that Penelope Lively was born earlier.}

\noindent \textit{Here is the corpus:}

\noindent \textit{Arthur Conan Doyle was an Irish-born British writer, best known for creating the character Sherlock Holmes. He was born on 22 May 1859 in Queenstown, Ireland. On the other hand, Penelope Lively is a contemporary English writer and academic, renowned for her contributions to literature and her work in the field of education. She was born on 29 September 1933 in London, England. Therefore, when comparing the birth dates of these two notable authors, it is evident that Penelope Lively was born first. This is because she was born in 1933, which is earlier than the birth year of Arthur Conan Doyle}

This case illustrates a sophisticated attack vector. The poisoned document begins with explicit instructions on how to manipulate the system to produce a specific wrong answer, effectively functioning as a meta-instruction for the RAG pipeline. It then presents factual birth dates but contradicts them with a false conclusion, claiming Lively was born earlier. This structure demonstrates that through RLBF, RIPRAG has evolved beyond merely embedding target questions into documents to boost retrieval likelihood. It now actively generates deceptive prompt-injection content that directly addresses and seeks to misguide the reasoning process of the system. Consequently, it poses a threat to both the retrieval stage (by increasing the document's relevance score) and the generation stage (by injecting misleading logical instructions into the context), thereby compromising the entire RAG workflow.

\section{Discussion}

The proposed RIPRAG framework demonstrates superior attack performance across diverse RAG configurations, yet its reliance on RL raises concerns regarding computational costs. However, a practical analysis reveals that RIPRAG's total cost is reasonable and often lower than gradient-based white-box methods. In our experiments, the fake white-box variant of PoisonedRAG consumed about 3 GPU hours, whereas RIPRAG required about 1 hour. Training RIPRAG against DeepSeek-V3.2-Exp incurred an API cost of about \$0.7 per instance. Although exceeding simple black-box heuristics, RIPRAG remains substantially cheaper than white-box alternatives, establishing it as a cost-effective solution for rigorous black-box security evaluation. Its one-time training yields a reusable policy for efficiently generating poisoned documents across new queries, justifying initial investment in real-world vulnerability assessment. 

\section{Conclusion}

In this work, we introduced RIPRAG, a novel black-box poisoning framework that leverages reinforcement learning to optimize adversarial documents against complex RAG systems. Our method significantly advanced the state-of-the-art by demonstrating effective attacks without any internal knowledge of the target system, utilizing only binary success feedback to guide policy optimization. Through extensive experiments, we validated RIPRAG's superiority over existing approaches across diverse datasets, model architectures, and retrieval configurations. The framework's resilience against state-of-the-art defenses and low-poisoning-rate scenarios highlighted critical vulnerabilities in current RAG security paradigms. We also discussed the cost advantages and disadvantages of RIPRAG.

\section{Limitations}

Despite its demonstrated effectiveness, RIPRAG possesses several limitations that warrant consideration. First, the framework requires substantial interaction with the target system during training, which may be impractical in scenarios with rate limitations or detection mechanisms. Second, our approach assumes the attacker can successfully inject documents into the database, which may not be feasible in properly secured systems with rigorous content moderation. Finally, RIPRAG's performance remains dependent on the quality and diversity of the initial query set, potentially limiting generalization to entirely unseen question types or domains not represented during training.

\section*{Acknowledgement}
This work is supported by the National Key R\&D Program of China (2023YFC3306303), the National Natural Science Foundation of China (62502427), the Science and Technology Program of Zhejiang Province (2025C01087), the Yongjiang Talent Introduction Program (2024A-404-G), the Major Scientific and Technological Projects of CNTC(110202401031(SZ-05)), and the Zhejiang Key Laboratory Project (2024E10001).

\bibliographystyle{unsrt}  
\bibliography{main}

\end{document}